\def\year{2022}\relax
\documentclass[letterpaper]{article} % DO NOT CHANGE THIS
\usepackage{aaai22}  % DO NOT CHANGE THIS
\usepackage{times}  % DO NOT CHANGE THIS
\usepackage{helvet}  % DO NOT CHANGE THIS
\usepackage{courier}  % DO NOT CHANGE THIS
\usepackage[hyphens]{url}  % DO NOT CHANGE THIS
\usepackage{graphicx} % DO NOT CHANGE THIS
\usepackage{natbib}  % DO NOT CHANGE THIS AND DO NOT ADD ANY OPTIONS TO IT
\usepackage{caption} % DO NOT CHANGE THIS AND DO NOT ADD ANY OPTIONS TO IT
\DeclareCaptionStyle{ruled}{labelfont=normalfont,labelsep=colon,strut=off} % DO NOT CHANGE THIS
\usepackage{algorithm}
\usepackage{algorithmic}

\usepackage{makecell}%表格内部换行
\usepackage{color}%字体颜色设置
\usepackage{graphicx}%表格竖排
\usepackage{newfloat}
\usepackage{listings}
\floatstyle{ruled}
\newfloat{listing}{tb}{lst}{}
\floatname{listing}{Listing}

\title{Impact of Scaled Image on Robustness of Deep Neural Networks}
\author {
    Chengyin Hu \textsuperscript{\rm 1},
    Weiwen Shi \textsuperscript{\rm 1}
}
\affiliations {
    \textsuperscript{\rm 1} University of Electronic Science and Technology of China\\
    cyhuuestc@gmail.com, Weiwen\_shi@foxmail.com
}

\usepackage{bibentry}
\begin{document}

\maketitle

\begin{abstract}
Deep neural networks (DNNs) have been widely used in computer vision tasks like image classification, object detection and segmentation. Whereas recent studies have shown their vulnerability to manual digital perturbations or distortion in the input images. The accuracy of advanced DNNs are remarkably influenced by the data distribution of their training dataset. Scaling the raw images creates out-of-distribution data, which makes it a possible adversarial attack to fool the networks. In this work, we propose a Scaling-distortion dataset ImageNet-CS by Scaling a subset of the ImageNet Challenge dataset by different multiples. The aim of our work is to study the impact of scaled images on the performance of advanced DNNs. We perform experiments on several state-of-the-art deep neural network architectures on the proposed ImageNet-CS, and the results show a significant positive correlation between scaling size and accuracy decline. Moreover, based on ResNet50 architecture, we demonstrate some tests on the performance of recent proposed robust training techniques and strategies like Augmix, Revisiting and Normalizer Free on our proposed ImageNet-CS. Experiment results have shown that these robust training techniques can improve networks’ robustness to scaling transformation.
\end{abstract}

\section{Introduction}

Nowadays, there is no surprise to see deep neural networks perform well on computer vision applications, such as image classification and object detection. Due to their high accuracy and excellent scalability on large-scale datasets, they are widely used in day-to-day life. However, recent studies have revealed the intrinsic vulnerability to small additive distortion or perturbations in the input data, which raises beating concerns about the robustness and security of deep neural networks. To ameliorate these issues, some researchers try to find more robust networks architectures \cite{ref44,ref46}, while the others commit to studying more robust training strategies \cite{ref49,ref50}.

The development of neural networks’ architecture is the foundation for robustness related research. Krizhevsky et al. \cite{ref1} proposed AlexNet, one of the first deep neural networks using convolutional layer to perform feature extraction, which won the first place in the 2012 Imagenet \cite{ref2} image classification competition, surpassing many traditional algorithms. The success of AlexNet started a boom in deep neural networks, motivating subsequent scholars to make more improvements to the architecture of neural network, and improve the networks’ accuracy in image classification. Based on AlexNet, VGG \cite{ref21} replaced the convolutional kernels with smaller ones, which reduces the computational cost and improves the generalization capability of the network with the same receptive field. GoogleNet \cite{ref22} proposed a module called inception to simulate sparse networks with dense construction. ResNet \cite{ref30} proposed residual modules to reduce the difficulty of learning identity maps, thus solving deep networks’ degradation problem.

Recently, many researchers have studied the vulnerability of deep neural networks \cite{ref3,ref4,ref10}. It has been known that many types of perturbations of the input data can change the output label of the network, such as random noises and adversarial perturbations \cite{ref5,ref6,ref7}. However, most of above works have mainly focused on the impact of local, small, and imperceptible perturbations on the classification results, while few studies have been conducted on the impact of global, geometric and structure transformation of the input. As a result, we have studied the impact of Scaling of the input images on the models’ output. To further investigate the impact of the Scaling of digital images in deep neural networks, we propose an image dataset with different scaling multiples generated from a subset of the Imagenet challenge dataset. It is well known that the performance of deep networks is highly relevant to the data distribution of its training dataset. Whereas scaling of the image produces out-of-distribution data, which misleads the networks into making incorrect predictions. Given the universality of image scaling in daily life, this deficit may have serious effects on the security of neural networks. For example, in the context of self-driving cars, it is fundamental to accurately recognize cars, traffic signs, and pedestrians, with these objects naturally scaling in the vehicles’ camera.

\begin{table*}[htbp]
    \centering
    \caption{\label{Table 1} Input sizes of well-known DNNs.}
    \begin{tabular}{cc}
    \hline
    Model & Size ($Pixel\times Pixel$)\\
    \hline
    LeNet-5 \cite{ref28} & $32\times32$\\
    \hline
    AlexNet \cite{ref1} & $227\times227$\\
    \hline
    VGG16 \cite{ref21},\\ ResNet \cite{ref30},\\ GoogleNet \cite{ref22} & $224\times224$\\
    \hline
    Inception V3 \cite{ref31} \cite{ref28} & $229\times229$\\
    \hline
    DAVE-2 \cite{ref29} & $200\times66$\\
    \hline
    \end{tabular}
\end{table*}

\begin{figure}
\centering
\includegraphics[width=1\linewidth]{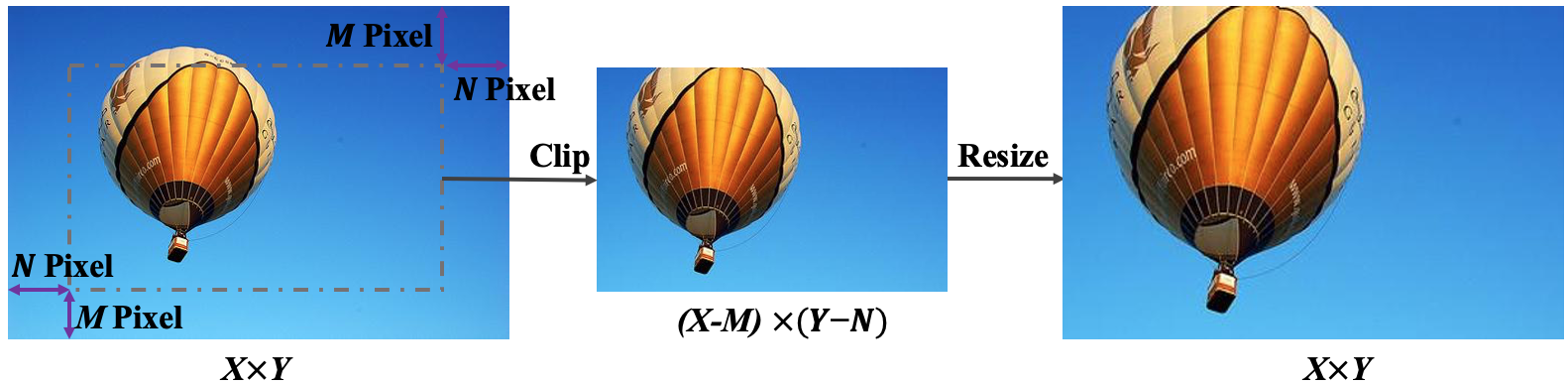} 
\caption{Generating a scaled sample.}
\label{figure1}
\end{figure}

Scaling of the images is one of the most common data augmentation techniques \cite{ref8,ref9,ref11,ref12}. By performing data augmentation, neural networks can be prevented from learning irrelevant features, fundamentally improving the overall performance \cite{ref13,ref14,ref20}. But to the best of our knowledge, there is little known about how deep neural networks process scaling transformation. Recent work by Xiao et al. \cite{ref34} proposed a scaling camouflage attacks, successfully attacking against famous cloud-base image services. The current robustness benchmarking datasets like Imagenet-C, providing out-of distribution cases related to noise, blur, weather, cartoons, sketches, etc. Hendrycks et al. \cite{ref16} and Lau F \cite{ref18} respectively proposed a challenging dataset, Imagenet-A and NAO, which consist of real-world unmodified natural adversarial examples that most famous deep neural networks fail. To the best of our knowledge, there is no dataset dedicated to render scaling images and scaling information to understand the behavior of deep neural networks.

The main contributions of this paper include the creation of a dataset related to scaling images to understand their impact on the task of classification and then analyze the performance of most famous deep network architectures on image classification task on the proposed dataset under different scaling multiples based on the classification accuracy. Finally, we used ResNet50 as an example to investigate the effect of robust learning strategies like Augmix \cite{ref44}, Revisiting \cite{ref45} and Normalizer Free \cite{ref46} on networks’ capability to abate the impact of scaling. The rest of the paper is organized as follows: Section 2 presents background information related to the existing literature, and Section 3 presents how we constructed the dataset and provides details of the experiments, followed by the results and findings in Section 4 and finally the conclusion and some discussion are given in Section 5.

\section{Background}
The impact of training data of neural networks on their performance have been a hot topic since they were proposed. In this section we present a review of relevant literature as well as some background knowledge.

The currently accepted hypothesis is that neural networks learn feature representation as well as semantic information in the data distribution of their training datasets. However, when the images are affected by perturbations like geometric transformation, deletion, and blur, their data distribution will change, which probably leading to the neural networks misclassification. Dodge and Karam \cite{ref24} checked how distortion and perturbations affect the classification paradigm of a Deep Neural Networks, and provide a brief outlook on how adversarial samples affect the performance of the DNN. They performed experiments on the ImageNet dataset with several common networks (Caffe Reference \cite{ref19}, VGG16 \cite{ref21} and GoogLeNet \cite{ref22}). Their test was on a subset of the validation set of the ImageNet. The results indicate that the deep neural networks are influenced by distortions, especially noise and blur. Dodge and Karam \cite{ref23} further compare the ability between human and deep neural networks to classify distorted images, and found that humans outperform neural networks on distorted stimuli, even when the networks are retrained with distorted data. Borkar and Karam \cite{ref24} evaluate the effect of perturbations like Gaussian blur and additive noise on the activations of pre-trained convolutional filters. They further propose a criteria to rank the most noise vulnerable convolutional filters in order to gain the highest improvement in classification accuracy upon correction. Zhou et al. \cite{ref27} showed fine-tuning and re-training would improve the performance of deep networks when classifying distorted images. Hossain et al. \cite{ref26} analyzed the performance of VGG16 when influenced by Gaussian white noise, Scaling Gaussian noise, salt \& pepper noise, speckle, motion blur, Gaussian blur. They used a training strategy called discrete cosine transform to improve the robustness when facing above distortions.

\begin{figure*}
\centering
\includegraphics[width=1\linewidth]{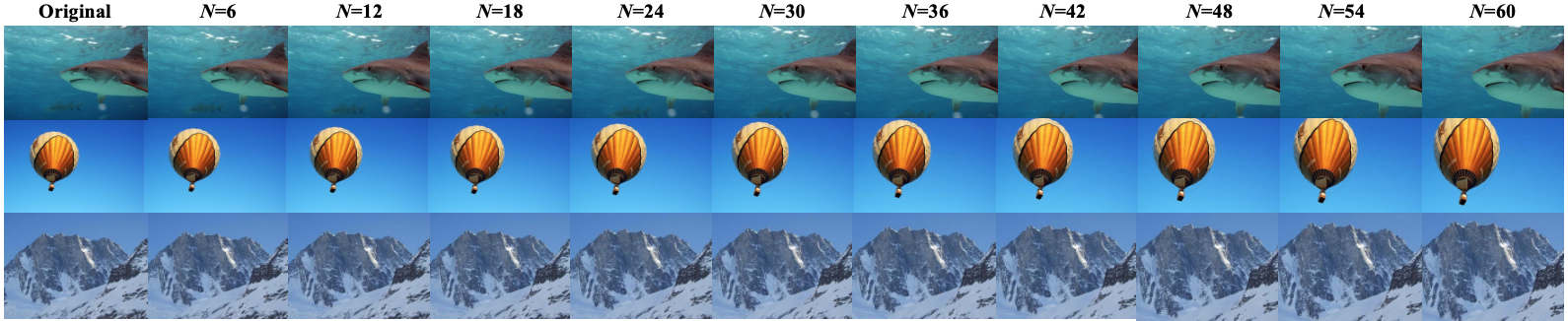} 
\caption{Example of images from the Imagenet-CS dataset.}
\label{figure2}
\end{figure*}

\textbf{Impact of Scaling.} Image scaling refers to changing the size of a digital image while preserving its visual features or visual semantics. Image scaling is actually a common action in deep learning applications. For simplicity and efficiency, a deep learning neural network model usually requires a fixed input scale. For image classification models, input images are usually resized to less than $300\times300$ to ensure high-speed training and classification. As shown in Table \ref{Table 1}, we show seven well-known deep neural networks and all of them use a fixed input scale for their training and classification process. Image scaling can be divided into upscaling and downscaling, depending on the size of the image is enlarged or reduced. When upscale an image, it is necessary to select a interpolation algorithm, which refers to inferring the pixel of the upscaled image according to the pixel values of the known image. The most commonly used interpolation algorithms are nearest-neighbor, bicubic, and bilinear \cite{ref28}. Formally, each interpolated point P can be regarded as the weighted average of the known points Q.

\begin{equation}
    \label{Formula 1}
    P=[{a}_{1},{a}_{2},...,{a}_{n}]{[{Q}_{1},{Q}_{2},...,{Q}_{n}]}^{T}
\end{equation}

where $[{a}_{1},{a}_{2},...,{a}_{n}]$ are the weights vectors for the known points.

The process of downscaling is the inverse process of the upscaling. Similarly, a certain downsampling algorithm need to be used to select some pixels of the known images to generate downscaled ones. Despite much work investigating the impact of image quality on deep neural networks, little work is done on effects of scaling. Sharma et al. \cite{ref33} were first analyzed the effect of different image scaling algorithms in face recognition applications. Xiao et al. \cite{ref34} have recently presented the vulnerability of deep learning models to the image scaling, and proposed a automated scaling attack algorithm. Quiring et al. \cite{ref35} analyzed the scaling attacks from the perspective of signal processing and identify their root cause as the interplay of downsampling and convolution. ZHENG et al. \cite{ref36} proposed a metric to measure the impact of image scaling on the robustness of adversarial examples and perform experiments on VGG-11 and Inception-v3. Kim et al. \cite{ref38} presents an image-scaling attack detection framework to detect image scaling attack and decline the impact of it. These results give us the motivation to further explore the impact of scaling and scaling related distortions on image classification. Based on the work from Hendrycks et al. \cite{ref39} we also used the validation set of the Imagenet dataset as our baseline and augmented different scaling images from these images. The details of the dataset generation are explained in Section 3.1. 

\textbf{Architectures.} The evolution of different architectures of deep neural networks began with Alexnet. It combines convolutional layers, pooling layers, activation functions and used GPU to accelerate the computation. The success of AlexNet on Imagenet has sparked enthusiasm for neural networks. VGG took advantage of Alexnet and made some improvement. It used several consecutive 3x3 convolution kernels instead of the larger ones in AlexNet ($11\times11$, $7\times7$, $5\times5$). For a given receptive field, i.e. the size of the region in the input that produces the feature, using stacked smaller convolution kernels is better than a larger convolution kernel, because it makes the network deeper and more efficient. With the development of the architecture of deep networks, the models became much deeper, which caused vanishing gradients and degradation problems. To tackle these problems, the most important innovations was the ResNet architecture which is still one of the widely used backbones in computer vision tasks. The ResNet proposed a structure called residual block, which skip connections between adjacent layers, enabling the network to learn identity mapping easier. It ensures that the deeper network at least perform as good as smaller ones. Another key innovation in deep network architectures is the inception module, which computes $1\times1$, $3\times3$ and $\times$ convolutions within the same module of the network, helping to learn a better representation of the image. Another significant architecture is the DenseNet \cite{ref48}. Based on ResNet architecture, DenseNet proposes a more radical intensive connectivity mechanism, which connects all the layers to each other and each layer takes all the layers before it as its input. DenseNet needs fewer parameters compared to the other traditional convolutional neural networks by reducing the need to learn redundant features. One of the drawbacks of deep neural networks is that they are computationally intensive and the models require a lot of memory which makes them unsuitable for mobile devices. To deploy models in such devices a group of lightweight networks were proposed, and Mobilenet \cite{ref40} is one of their best-known representatives. It proposed the concepts of depth wise separable convolutions and inverted residuals, which achieve similar performance to traditional networks with less computational cost.

\begin{figure*}
\centering
\includegraphics[width=1\linewidth]{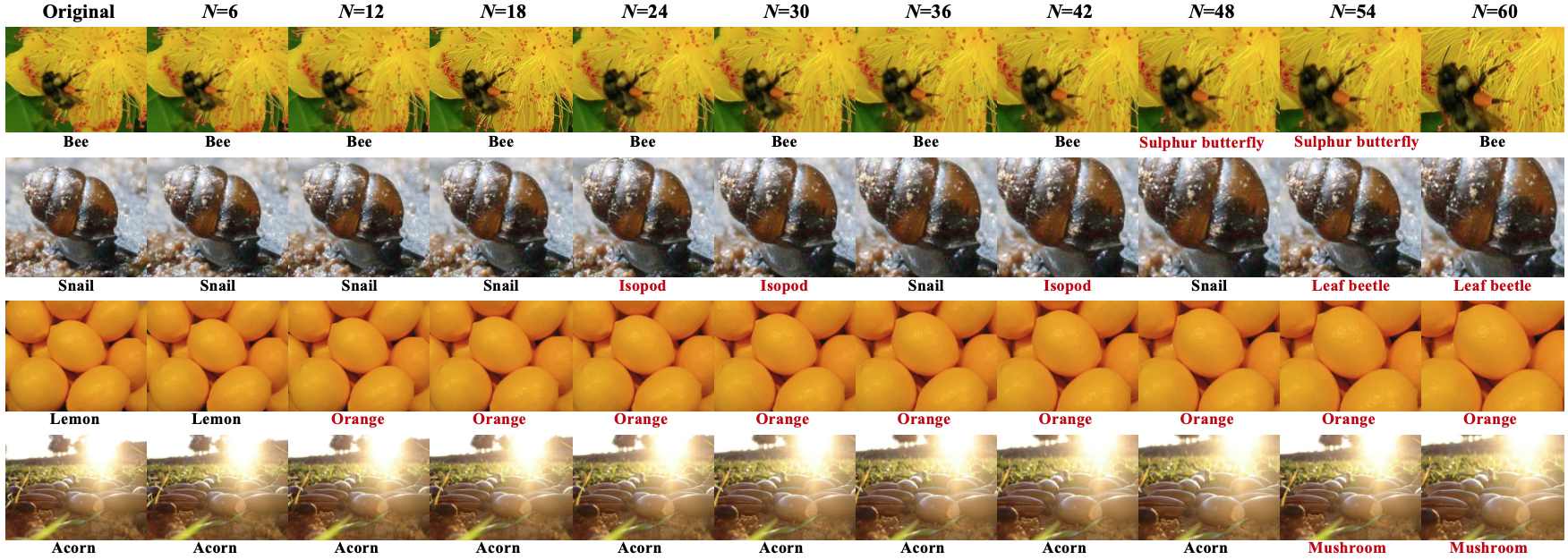} 
\caption{Some misclassifications of ResNet50.}
\label{figure3}
\end{figure*}

\textbf{Robustness.} Evaluating robustness of deep neural networks is still a challenging and ongoing area of research. Papernot \cite{ref41} et al. first pointed out some limitations of deep learning in adversarial settings. They proposed forward derivative attack to fool deep networks by only alter a minority of input. Hendrycks et al. \cite{ref42} defined some benchmark metrics of robustness of deep neural network to some common perturbations like additive noise, blur, compression artifacts, etc. They proposed a variant of Imagenet referred to as Imagenet Challenge (or Imagenet-C). Imagenet-C contains 15 types of automated generated perturbations, on which many well-known deep networks perform poorly. Kanjar et al \cite{ref17}. analyzed the impact of color on robustness of widely used deep networks. Recent studies have indicated that deep convolutional neural networks pre-trained on Imagenet dataset are vulnerable to texture bias \cite{ref43}, while the impact of scaling in images is not deeply studied. Xiao et al. \cite{ref34} have formalized the scaling attack, illustrating its goal, generation algorithms, and optimization solution. Zheng et al. \cite{ref36} provide a basis for robustness evaluation and conduct experiments in different situations to explore the relationship between image scaling and the robustness of adversarial examples. While the adversarial attack techniques are developing, many studies focus on defense against these attacks and try to find feasible training strategies to improve the robustness of models \cite{ref11}. Augmix \cite{ref44} is a simple training strategy which uses several augmentation techniques together with Jenson-Shannon divergence loss to enforce a common embedding for the classifier. Brock et al. \cite{ref46} proposed a normalized family of free networks called NF-Nets to prevent the gradient explosion by not using batch normalization. Tan et al. \cite{ref45} recently showed that network models can effectively improve the classification performance of ResNet models by using some scaling strategies and developed a set of models called ResNet-RS.

\section{Experiments and Methodology}
In this section, we present the details of data generation process of our proposed dataset and the neural network architecture used in our experiments.

\subsection{Dataset Generation}
Typically, DNNs is trained on an Imagenet dataset of 1000 categories. Our proposed dataset is derived from Imagenet Challenge with several scaling operations, thus we call it Imagenet-Challenge-Scaling (Imagenet-CS). Firstly, 50 images were randomly selected from each category of Imagenet to generate a clean sample dataset, which totally contains 50,000 raw images. Secondly, the Imagenet-CS data set of 500,000 images is obtained by 10 different degrees of magnification for each image. As shown in Figure \ref{figure1}, the method to generate scaled image is: (1) Clip the picture to $(X - M) \times (Y - N)$ pixels, where $X$ and $Y$ are the length and width of the raw image, while $M$ and $N$ are size of pixels clipped in the long and wide directions respectively. (2) Upscale the picture to its original size with bilinear interpolation algorithm.

For simplicity, we set the clipped pixel values M = N, and switch their values from 6 to 60 pixels, spaced 6 apart. As the magnification increases, it is obvious that the image will be only part of the object, while its semantic information remains unchanged. Figure \ref{figure2} shows some samples of Imagenet-CS. In next section, the performance of advanced DNNs on the scaled images with different magnification will be shown.

\subsection{Impact on widely used network architectures}

In this section, we perform some experiments on the accuracy of six widely used depp network architectures mentioned in Table \ref{Table 2} on the generated Imagenet-CS dataset. To establish a reference, the first column (Original) represents the classification accuracy of the images in the absence of any scaling. 

\begin{table*}[htbp]
    \centering
    \caption{\label{Table 2} Classification Top-1 accuracy (\%) of well-known networks on the Imagenet-CS dataset.}
    \begin{tabular}{cccccccccccc}
    % \hline
    $N$ & $0$ & $6$ & $12$ & $18$ & $24$ & $30$ & $36$ & $42$ & $48$ & $54$ & $60$\\
    \hline
    DenseNet & 82.14 & 80.91 & 80.53 & 80.06 & 79.32 & 78.54 & 77.45 & 76.35 & 75.00 & 73.50 & 72.31\\
    \hline
    ResNet & 85.83 & 84.60 & 83.94 & 88.33 & 82.58 & 81.61 & 80.72 & 79.49 & 78.14 & 76.50 & 74.97\\
    \hline
    VGG19 & 82.51 & 81.08 & 80.38 & 79.62 & 78.71 & 77.55 & 76.09 & 74.71 & 73.14 & 71.19 & 69.46\\
    \hline
    GoogleNet & 75.87 & 75.04 & 74.69 & 74.10 & 73.56 & 72.69 & 71.65 & 70.45 & 69.14 & 67.85 & 66.33\\
    \hline
    MobileNet & 80.70 & 79.39 & 78.97 & 78.36 & 77.59 & 76.71 & 75.62 & 74.26 & 72.93 & 71.31 & 69.75\\
    \hline
    AlexNet & 71.59 & 70.51 & 70.06 & 69.28 & 68.30 & 67.09 & 65.79 & 63.99 & 62.28 & 60.47 & 58.62\\
    \hline
    \end{tabular}
\end{table*}

As can be seen from the experimental results in Table \ref{Table 2}, with the increase of image upscaling, the classification accuracy of networks is decreasing. This observation is consistent throughout all architectures. When further investigate the influence of different architectures, it is found that the ResNet architectures perform best among these models. DenseNet and VGG19 architectures show similar classification accuracy and perform better than MobileNet and GoogleNet. And AlexNet show the worst classification performance in every magnification. 

This phenomenon indicates that although the semantic features of the images are barely changed, the the scaling operation on images still have significant impact on the performance deep neural networks. Despise the tested deep networks all contains convolutional layers, which gives them the property of geometric invariance, they are still vulnerable to scaling transformation. The process of scaling the image can be regarded as a decrease of the distance between the photographer and the object, which means that when the distance of the image is changed, the classifier will probably make a misclassification. Figure \ref{figure3} shows a example that ResNet50 misclassify images as they were scaled. For example, Snail is misclassified as Isopod, Leaf beetle, etc. Meanwhile, it can be seen from Figure 3 that when the image is slightly enlarged, the semantic information of the image does not change, but the classifier makes a wrong classification judgment, which explains that the current advanced model is trained on a relatively single dataset (for example, pictures taken at the same distance). On the other hand, it can be seen from the classification of Fugure 3 Snail that the advanced classifier shows discontinuous misclassification rather than continuous misclassification. This means that the classification boundary of the model is relatively dense.

\subsection{Recent advances in efficient and robust models}

\begin{figure}
\centering
\includegraphics[width=1\linewidth]{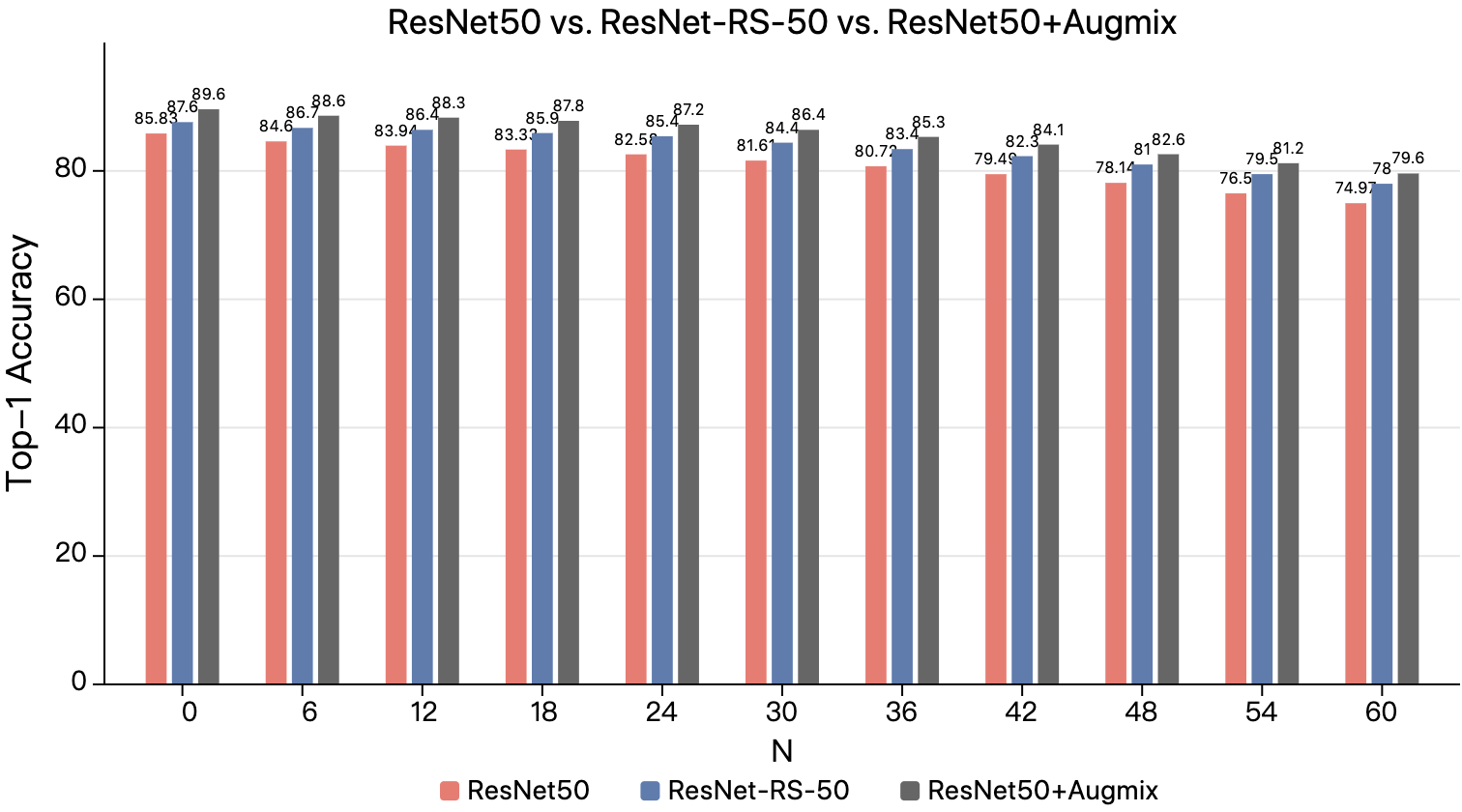} 
\caption{Performance of ResNet50 vs. ResNet-RS-50 vs. ResNet50+Augmix.}
\label{figure4}
\end{figure}

\textbf{Augmix and ResNet-RS-50.} Augmix \cite{ref44} is a data processing technology used to improve the robust performance of DNNs, including rotation, translation, separation and other enhancement technologies. It achieves simple data processing within limited computational overhead, helping the model withstand unforeseen corruptions, and AugMix significantly improves robustness and uncertainty in challenging image classification benchmarks. Here we show the classification performance of ResNet50 with Augmix data processing technology on Imagenet-CS. We compared the classification performance of pretrained ResNet50 model with that of pretrained ResNet50 model with Augmix technology, and the classification comparison is shown in Figure \ref{figure4}. It can be seen that on Imagenet-CS, the classification performance of ResNet50 with Augmix technology is better than that of ordinary pretrained ResNet50 model. Bello et al. \cite{ref45} recently showed that scaling network models can effectively improve the classification performance of models, and developed a set of models called ResNet-RS. It pointed out that the training and extension strategy may be more important than the architecture changes, and the ResNet architecture designed with the improved training and extension strategy is 1.7-2.7 times faster than the EfficientNets on TPUs, while achieving similar accuracy on ImageNet. In the large-scale semi-supervised learning setting, ResNet-RS achieves 86.2\% Top-1 ImageNet accuracy while being 4.7 times faster than EfficientNet-NoisyStudent. Here, we show the classification performance of ResNet-RS-50 on Imagenet-CS and its Top-1 accuracy is shown in Figure \ref{figure4}. It can be seen that ResNet-RS-50 has better robustness than pretrained ResNet50.

In general, the robustness of ResNet50 with Augmix and ResNet-RS-50 are better than the original pretrained ResNet50 model. In addition, it can be seen from Figure \ref{figure4} that, as the image continues to be magnified, the classification accuracy of ResNet50 with Augmix and ResNet-RS-50 also gradually decreased, indicating that scaling still has certain antagonism to the improved ResNet50.

% Brock et al. \cite{ref46} proposed a normalized family of free networks called NF-Nets, which do not use batch normalization. In the training of NF-Nets, the gradient size is limited to effectively prevent gradient explosion and training instability. Figure \ref{figure5} shows the Top-1 accuracy of NF-ResNet50 on Imagenet-CS. It can be seen that, compared with pretrained ResNet50 model, the classification performance of NF-ResNet50 has achieved a certain improvement. 

\begin{table*}[htbp]
    \centering
    \caption{\label{Table 3} Performance of ResNet models with different depths.}
    \begin{tabular}{cccccccccccc}
    % \hline
    $N$ & $0$ & $6$ & $12$ & $18$ & $24$ & $30$ & $36$ & $42$ & $48$ & $54$ & $60$\\
    \hline
    ResNet18 & 77.90 & 76.93 & 76.25 & 75.75 & 75.05 & 73.96 & 72.72 & 71.47 & 69.94 & 68.45 & 66.90\\
    \hline
    ResNet34 & 83.21 & 82.06 & 81.58 & 80.88 & 80.22 & 79.14 & 78.04 & 76.64 & 75.41 & 73.75 & 72.00\\
    \hline
    ResNet50 & 85.83 & 84.60 & 83.94 & 83.33 & 82.58 & 81.61 & 80.72 & 79.49 & 78.14 & 76.50 & 74.97\\
    \hline
    ResNet101 & 88.51 & 87.04 & 86.59 & 86.11 & 85.38 & 84.58 & 83.51 & 82.38 & 81.00 & 79.53 & 78.13\\
    \hline
    ResNet152 & 88.92 & 87.82 & 87.56 & 86.95 & 86.37 & 85.48 & 84.52 & 83.27 & 81.90 & 80.41 & 79.00\\
    \hline

    \end{tabular}
\end{table*}

\begin{figure}
\centering
\includegraphics[width=1\linewidth]{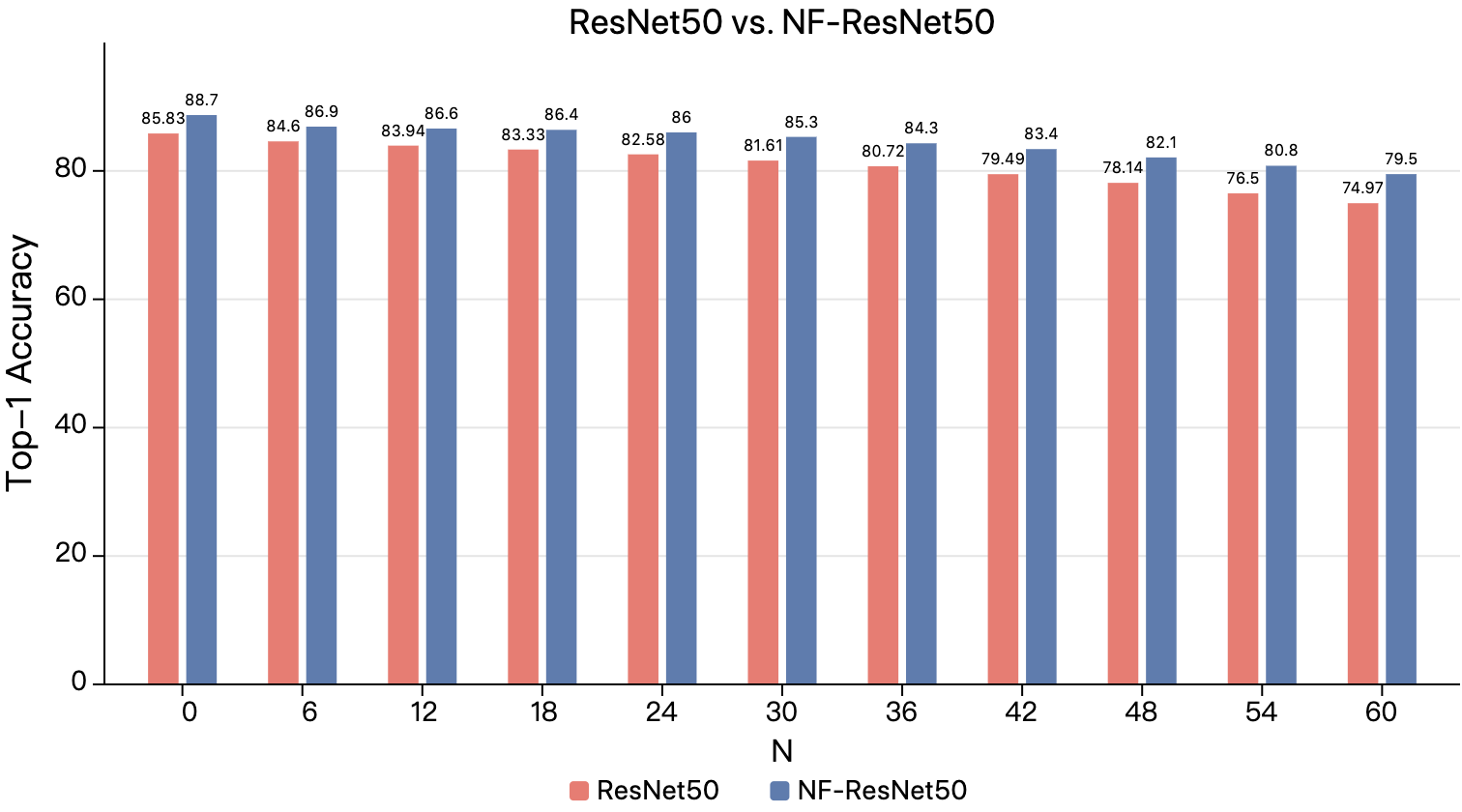} 
\caption{Performance of ResNet50 vs. NF-ResNet50.}
\label{figure5}
\end{figure}

\textbf{Normalizer Free ResNet50(NF-ResNet50).} Brock et al. \cite{ref46} abandoned the traditional concept that data need to be normalized, and proposed a deep learning model NF-Nets without normalization, which achieved the best level in the industry on large image classification tasks. They proposed Adaptive Gradient Clipping methods to realize non-normalized networks augmented with larger quantities of subscale and large scale data. In the training of NF-Nets, the gradient size is limited to effectively prevent gradient explosion and training instability. Figure \ref{figure5} shows the Top-1 accuracy of NF-ResNet50 on Imagenet-CS. It can be seen that, compared with pretrained ResNet50 model, the classification performance of NF-ResNet50 has achieved a certain improvement. 

\subsection{ResNet models with different depths}
Before ResNet, there were not many network layers. VGG's network has only 19 layers, but ResNet had a whopping 152 layers. Many people have the intuitive impression that the more layers of the network, the better the training effect, but in this case, why not use 152 layers instead of 19 layers in VGG network? In fact, the accuracy of the training model is not necessarily correlated with the number of model layers. Because with the deepening of the network layer, the network accuracy needs to appear saturation, will appear the phenomenon of decline.
Suppose a 56 network than 20 layer network training effect is poor, many people first reaction is fitting, but this is not the case, because the accuracy of fitting phenomenon of the training set will be very high, but 56 layer network training set accuracy may also is very low, it shows that the network depth increase may not guarantee the accuracy of classification. It is obvious that with the deepening of layers, there will be gradient disappearance or gradient explosion, which makes it difficult to train deep models. However, BatchNorm and other methods have been used to alleviate this problem. Therefore, how to solve the degradation problem of deep networks is the next direction of neural networks development.
ResNet is developed and optimized on the basis of AlexNet. One of the major advantages of residual neural network is identity mapping. The problem with AlexNet is that the optimization deteriorates as the number of layers increases. ResNet is designed to mitigate the vanishing/exploding gradient phenomenon by introducing residuals, but it's a mitigation, essentially. It's just that the valid path from loss to the input is shorter, and you can just add delta to the destination layer of shortcut when you take the derivative.

We use ResNet models with different depths (ResNet18, ResNet34, ResNet50, ResNet101, ResNet152) to test the classification top-1 accuracy on the proposed dataset. It can be seen from Table \ref{Table 3} that the classification accuracy of ResNet series models is proportional to the model depth. Similarly, the larger N is, the smaller is the classification top-1 accuracy of each model.

% \begin{figure}
% \centering
% % \setlength{\belowcaptionskip}{-0.5cm}
% \includegraphics[width=1\linewidth]{figures/fig6.png} 
% \caption{Performance of ResNet models with different depths.}
% \label{figure6}
% \end{figure}

\section{Conclusion}

With deep neural networks being widely used in our daily life, it is crucial to study the robustness of deep learning models and try to make these models more robust and accurate to perturbations. Experimental studies presented in this paper have yielded some interesting results with respect to the impact of scaling transformation of images on the performance of deep neural network architectures with respect to the shift in data distribution. The performance of these networks drastically reduces when the magnification of scaling increases. In this paper, we have presented the over all classification accuracy performance of some widely used deep neural network architectures under different scaling distortions and the interesting results demonstrated will serve as a motivation to investigate the scaling sensitivity of further architectures studies in the future. The analysis mentioned in this paper will motivate researchers to take into consideration the impact of scaling and aspects of other geometric transformation for proposing more accurate and robust models based on deep neural networks. The important observations are listed as follows:

\begin{itemize}

\item There is a significant impact of scaling transformation on the inference of deep neural networks. 
\item Data processing and data augmentation techniques like Augmix have some positive impact on robustness and optimizing the training procedure of deep networks, making Resnet RS-50 is much more robust model compared to Resnet-50 with respect to scaling. 
\item Training procedures like adversarial prop and noisy student training offer some amount of additional robustness to models. 
\item The Normalizer free models offer more robustness to scaling specific transformation. 

\end{itemize}

% Use \bibliography{yourbibfile} instead or the References section will not appear in your paper
% \nobibliography{aaai22}
\bibliography{aaai22}

\end{document}